\documentclass{article} 
\usepackage{iclr2024_conference,times}

\usepackage{amsmath,amsfonts,bm}

\def\eqref#1{equation~\ref{#1}}

\def\1{\bm{1}}

\def\vg{{\bm{g}}}

\DeclareMathAlphabet{\mathsfit}{\encodingdefault}{\sfdefault}{m}{sl}
\SetMathAlphabet{\mathsfit}{bold}{\encodingdefault}{\sfdefault}{bx}{n}

\usepackage{colortbl}
\usepackage{amsmath}
\usepackage{xcolor}
\usepackage{hyperref}
\usepackage{url}
\usepackage{graphicx}
\usepackage{subcaption}
\usepackage{todonotes}
\usepackage{booktabs}       
\usepackage{multirow}

\usepackage{CJKutf8}
\newcommand{\chinese}[1]{{\begin{CJK*}{UTF8}{gkai} #1 \end{CJK*}}}

\title{Sign2GPT: Leveraging Large Language Models for Gloss-Free Sign Language Translation}
\author{Ryan Wong$^1$, Necati Cihan Camgoz$^2$, Richard Bowden$^1$ \\
$^1$University of Surrey, 
$^2$Meta Reality Labs\\
\texttt {\{r.wong, r.bowden\}@surrey.ac.uk, neccam@meta.com}
}

\iclrfinalcopy 
\begin{document}

\maketitle

\begin{abstract}

Automatic Sign Language Translation requires the integration of both computer vision and natural language processing to effectively bridge the communication gap between sign and spoken languages. However, the deficiency in large-scale training data to support sign language translation means we need to leverage resources from spoken language. We introduce, Sign2GPT, a novel framework for sign language translation that utilizes large-scale pretrained vision and language models via lightweight adapters for gloss-free sign language translation. The lightweight adapters are crucial for sign language translation, due to the constraints imposed by limited dataset sizes and the computational requirements when training with long sign videos.
We also propose a novel pretraining strategy that directs our encoder to learn sign representations from automatically extracted pseudo-glosses without requiring gloss order information or annotations.
We evaluate our approach on two public benchmark sign language translation datasets, namely RWTH-PHOENIX-Weather 2014T and CSL-Daily, and improve on state-of-the-art gloss-free translation performance with a significant margin.

\end{abstract}

\section{Introduction}

Sign languages are the primary form of communication for millions of Deaf individuals. 
Sign languages make use of complex visual gestures as a form of communication \citep{braem2001hands}. 
Automatic sign language translation is a challenging task for both natural language processing and computer vision, as it requires understanding of both sign and spoken language semantics.

Many prior studies have heavily relied on gloss annotations as a means to achieve better translation performance \citep{camgoz2018neural,camgoz2020sign,ye2023cross,yao2023sign,zhang2023sltunet}, where gloss annotations represent written descriptions of signs.
These glosses are provided in sign order, aiding in the regularization of models to learn valuable sign features necessary for continuous sign language recognition and translation. 
However, creating datasets with gloss annotations is resource-intensive and time-consuming.
Hence, a recent trend is to shift towards gloss-free sign language translation \citep{camgoz2020multi,yin2023gloss,zhou2023gloss}, which is the primary focus of our paper. 
Gloss-free sign language translation poses significant challenges due to the unique grammatical structures and vocabulary of sign languages.
Limited data availability, individual signing style variation and long sign videos add to the complexity of the task.

\begin{figure*}[ht]
    \centering
    \includegraphics[width=1.0\textwidth]{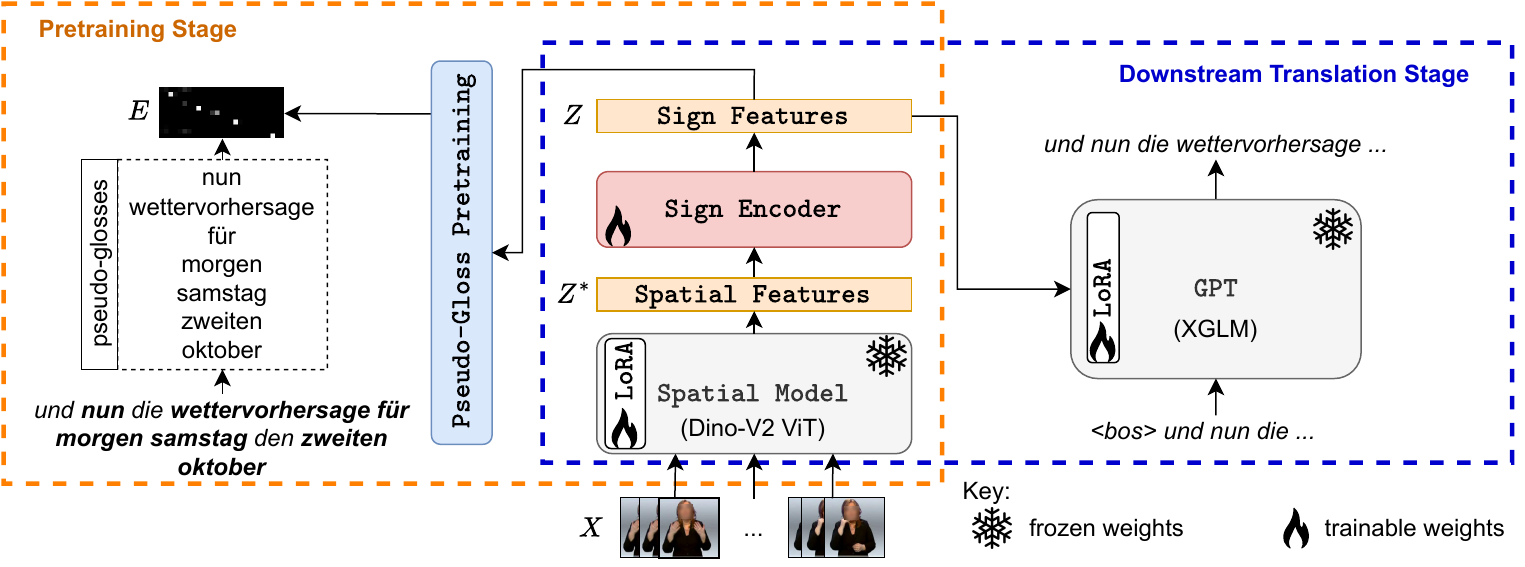}
    \caption{Overview of Sign2GPT, which consists of a pretraining stage that makes use of pseudo-glosses and downstream translation that leverages a frozen GPT model.}
    \label{fig:overview}
\vspace{-1.20em}
\end{figure*}

In this paper, we propose Sign2GPT for sign language translation to address the aforementioned challenges. 
As depicted in Figure~\ref{fig:overview}, Sign2GPT utilizes pretrained large vision \citep{oquab2023dinov2} and language models \citep{lin2021few}, benefiting from their extensive training on large-scale datasets to improve sign language translation performance. 
We also present a novel pretraining strategy before the translation task, which encompasses two key components. Firstly, we develop an algorithm designed to automatically generate pseudo-glosses. Secondly, we introduce a prototype driven method for pretraining the sign encoder using the pseudo-glosses. 
Notably, our approach eliminates the necessity of manual gloss annotations as well as glosses to be in sign order.

To combat over-fitting and address memory constraints resulting from long video sequences during training, we adopt a strategy of freezing the external pretrained models and utilize specialized low-rank adapters. Inspired by \citet{hu2021lora}, we employ low-rank adapters as a method for facilitating the adaptation of frozen Generative Pretrained Transformers (GPT) \citep{radford2018improving, lin2021few} as well as a Vision Transformer (ViT) \citep{dosovitskiy2020image, oquab2023dinov2} to the specialized domain of sign language. Our contributions can be summarized as follows: (1) We introduce an end-to-end gloss-free sign language model, Sign2GPT, designed for sign language translation, leveraging a frozen GPT language model, (2) We propose a novel pseudo-gloss pretraining strategy, utilizing automatically extracted pseudo-glosses from sentences to pretrain the sign encoder. (3) Sign2GPT demonstrates significant performance improvements over previous gloss-free sign language translation approaches, offering a promising pathway for adapting frozen language and vision models to the domain of sign language translation.
\vspace{-0.15em}

\section{Related Work}

Isolated Sign Recognition (ISR), the task of identifying individual signs within a sign video, has been extensively studied in the literature.  
Many ISR approaches leverage deep learning, adapting action recognition models \citep{carreira2017quo,tran2018closer,yan2018spatial,jiang2021skeleton,albanie2020bsl} on datasets such as WLASL \citep{li2020word}, MSASL \citep{joze2018ms} and AUTSL \citep{sincan2020autsl}. 
New approaches have explored the integration of linguistic priors to enhance the accuracy of ISR models \citep{zuo2023natural,wong2023learnt}.
This is achieved through the utilization of pretrained word embeddings \citep{joulin2016fasttext} as features to regularize the models during training.

While these methodologies have been successful for ISR, they address only a portion of the sign language translation challenge. 
Consequently, new datasets for Continuous Sign Language Recognition have been developed, such as  RWTH-PHOENIX-Weather 2014 \citep{koller2015continuous} and CSL-Daily \citep{zhou2021improving}. 
These datasets consist of sequences of coarticulated signs in a continuous video, accompanied by target labels known as glosses.
These glosses are arranged in sign order and serve as an intermediary representation bridging the gap between sign videos and spoken language translation.
Many approaches use Connectionist Temporal Classification (CTC) loss \citep{graves2006connectionist} to localize and recognize glosses within sign videos \citep{camgoz2018neural,camgoz2020sign,zhou2021improving,cui2017recurrent,zheng2023cvt} but this relies on the monotonicity of labels and content.

Automatic Sign Language Translation (SLT) is one of the main goals of computational sign language research, aimed at bridging the communication gap between sign languages and spoken languages. SLT datasets, such as RWTH-PHOENIX-Weather 2014T \citep{camgoz2018neural} and CSL-Daily \citep{zhou2021improving}, contain video sequences of signers and corresponding spoken language sentences. These videos are often much longer than action recognition datasets, which typically use a fixed-length short sequence as input \citep{soomro2012ucf101,carreira2017quo}.

SLT is generally divided into gloss-based and gloss-free translation. Gloss-based SLT involves translating sign language into spoken language with the aid of gloss annotations via the utilization of the aforementioned Continuous Sign Language Recognition methods. The gloss annotations are manually annotated by expert annotators and depict sign type and order. Initial approaches used CNN and RNN-based models \citep{camgoz2018neural}, but transformers have gained prominence in neural machine translation \citep{vaswani2017attention}, leading to the development of Transformer-based approaches for SLT \citep{camgoz2020sign,camgoz2020multi,chen2022two,gancontrastive,zhang2023sltunet}. Further improvements by progressive pretraining of large language models are also achieved by using manually annotated sign-ordered gloss as supervision \citep{chen2022simple}.

Recently, researchers have explored gloss-free sign language translation to avoid the dependencies of manually created gloss annotations. Approaches of finding glosses include using sign spotters \citep{sincan2023context,shi2022open,momeni2022automatic}, but these models are limited to the domain of the sign datasets and the vocabulary it was trained on. We take an alternative approach by automatically creating pseudo-glosses from the spoken language sentences. By using sentences and sign language priors, we automatically generate pseudo-glosses, overcoming limitations associated with pretrained sign spotting models and manually annotated labels.
While techniques for verifying the existence of words exist \citep{zhao2021conditional}, they utilize separate models for each vocabulary item and two-stage translation system, incurring high training costs. In contrast, we propose a singular model that efficiently determines both the localization and existence of pseudo-glosses. This model serves as the foundation for an end-to-end solution in sign language translation.

Other methods of gloss-free SLT make use of domain transfer through the integration of linguistically pretrained models to enhance SLT performance by using pretrained encoder-decoder models like mBart \citep{liu2020multilingual} and T5 \citep{raffel2020exploring} models to improve translation quality \citep{zhou2023gloss,uthus2023youtube}. We instead use a frozen Generative Pretrained Transformer (GPT) \citep{lin2021few,radford2018improving} as the decoder to enhance translation performance, rather than fine-tuning or progressively pretraining encoder-decoder language models used in previous approaches. 

Recently Low-Rank Adapters \citep{hu2021lora}, which update only a small number of weights, have been proposed to adapt frozen LLMs to spoken language tasks. They have also been applied to multi-modal learning for general visual understanding related tasks \citep{zhang2023llama,gao2023llama,zhang2023video}. These approaches use features extracted from CLIP \citep{radford2021learning} which are not suitable as a sign language representation due to being trained on image captions. Our proposed pretraining approach focuses on visual-linguistic sign language features and subsequently adapting them to spoken language models, harnessing the capabilities of LLMs.

\section{Method}

Our approach aims to leverage the linguistic knowledge inherent in LLMs to enhance Sign Language Translation (SLT). We address two main challenges: handling memory-intensive sign language videos with numerous frames and creating sign representations that seamlessly integrate with LLMs. In the following subsections, we outline our model architecture (Figure \ref{fig:overview}) and training strategy to tackle these challenges and enhance SLT using LLMs.

\subsection{Model Architecture}

\begin{figure*}[ht]
    \centering
    \includegraphics[trim={0 0.75em 0 0.7em},clip,width=0.65\textwidth]{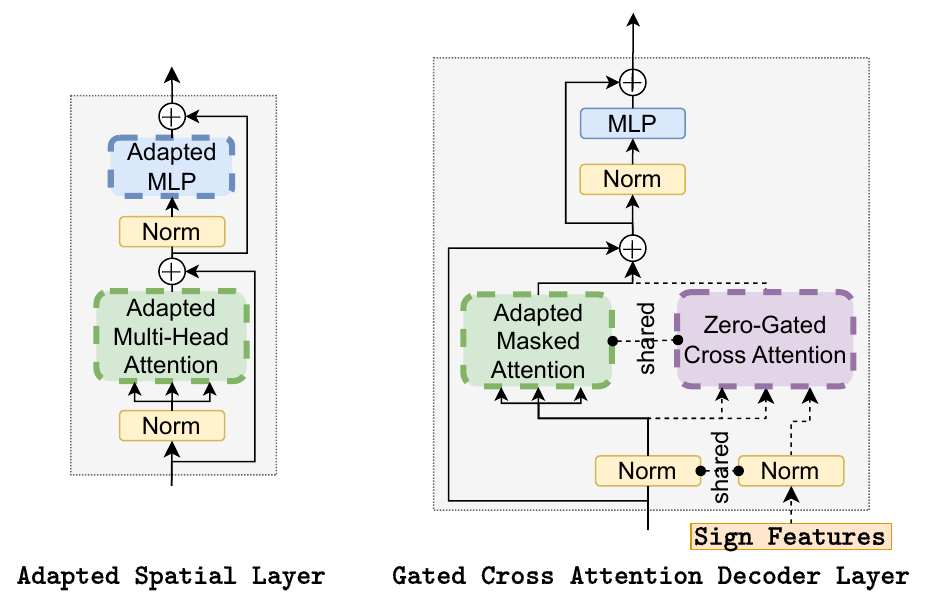}
    \caption{Overview of adapting layers in the spatial model layers (left) and decoder layer (right). We make use of adapters that introduce new low-rank weights to blocks shown by the dashed lines while keeping the original pretrained weights frozen.}
    \label{fig:components}
\vspace{-1.0em}
\end{figure*}

\paragraph{Spatial Backbone.}
Our video-based sign language translation framework relies on the spatial model, designed to extract spatial features, $Z^*$, from input video frames denoted as $X = \{x_0, x_1, ..., x_{T^{*}}\}$ with $T^{*}$ frames. These features of dimension $C$, represented as $Z^* \in \mathbb{R}^{T^{*} \times C}$, are subsequently used by the sign encoder.
We employ the Dino-V2 Vision Transformer \citep{oquab2023dinov2}, specifically the ViT-S/14 variant. Dino-V2 is a self-supervised vision model known for its robust feature extraction capabilities across various visual tasks, making it a suitable choice for our framework. Fine-tuning Dino-V2 is essential for adapting it to the unique characteristics of sign language translation datasets.

The spatial model often involves a significant number of parameters, posing memory and computational challenges. To address this, we adopt LoRA (Low-Rank Adapters), a lightweight adaptation technique. 
LoRA has been shown to be effective in finetuning LLMs \citep{hu2021lora} and Diffusion models \citep{roich2022pivotal,ruiz2022dreambooth,gal2022image}.
LoRA is applied to the top encoder layers, targeting fully connected layers in the MLP and Multi-Head Attention as shown in Figure \ref{fig:components} (left).
We utilize the class token's output as a feature vector, undergoing linear transformation and batch normalization to produce $Z^* \in \mathbb{R}^{T^* \times C}$, feeding it into our sign encoder.

\paragraph{Sign Encoder.}
We utilize spatial features, $Z^*$,  as input to our sign encoder, aiming to learn spatio-temporal sign representations. Our translation model must handle sequences often comprising hundreds of frames. To meet this requirement, we designed a spatio-temporal transformer model, drawing inspiration from prior sign language translation approaches \citep{camgoz2020sign, yin2023gloss}. This model incorporates two crucial modifications to enhance efficiency and effectiveness.

Firstly, to address the challenge of processing many frames, we employ temporal downsampling after specific layers within our encoder. This downsampling reduces the temporal dimension from $T^*$ to $\frac{T^*}{2}$ using strided averaging with a kernel size of three and a stride of two. This design was chosen to balance between computational efficiency and the preservation of temporal information.

Secondly, we use local self-attention with a window size of seven, a technique proven to be highly effective in SLT tasks \citep{yin2023gloss}. This local attention mechanism is integrated with the temporal downsampling, extending the model's temporal receptive field deeper into the network. This streamlined approach not only conserves memory but also minimizes redundancy, making it highly compatible with the subsequent decoder model.
The output sign representation is denoted as $Z\in\mathbb{R}^{T\times C}$, with $T$ representing the output temporal dimension which has been downsampled where $T=\frac{T^*}{2}$.

\paragraph{Language Decoder.}
In the decoder, we adapt the XGLM model \citep{lin2021few}, a multilingual GPT Language Model known for its versatility in few-shot learning on text data. We chose the 1.7B parameter variant of XGLM to balance performance and memory utilization.
We draw inspiration from successful language model adaptation techniques like zero-gated cross-attention and LoRA which have been used to adapt LLMs to different textual and multi-model tasks \citep{gao2023llama, zhang2023llama, zhang2023video, alayrac2022flamingo}. 

Before passing the sign features to the decoder, we map our sign features to the decoder's dimension with a linear layer $\texttt{FC}_m$. To adapt the XGLM decoder for sign language translation, we employ the zero-gated multi-head cross-attention to the decoder layer as shown in Figure \ref{fig:components} (right). This enhancement shares weights from the pretrained masked multi-head attention and integrates a separate LoRA for masked multi-head attention (Adapted Masked Attention) and cross-attention (Zero-Gated Cross Attention). The sign features are first passed through the frozen decoder's layer normalization and then used as keys for cross-attention.
To integrate sign features without overshadowing linguistic features, we employ gated scaled dot-product attention into the cross-attention:
\begin{equation}
\text{GatedAttention}(Q, K, V) = \left( \vg \times \texttt{softmax}\left(\frac{QK^T}{\sqrt{d_k}}\right)\right) V
\end{equation}
K and V represent the inputs from the key and value derived from the sign features, while Q originates from the textual features. $\vg$ is a learnable gate parameter for each attention head which is clamped between 0 and 1 and initialized to zero to preserve linguistic knowledge at the start of training.
This modification allows XGLM to adapt seamlessly to sign language. The gate parameter and LoRA parameters drive this adaptability, gradually incorporating sign features while leveraging linguistic knowledge. As depicted in Figure~\ref{fig:components} (right), the outputs from the Adapted Masked Attention and Zero-Gated Cross Attention are summed together.

\subsection{Training Strategy}

Our model architecture (Figure \ref{fig:overview}) allows direct video-to-text training for SLT. We prioritize high-quality sign features by concentrating most of the trainable parameters in the sign encoder. We freeze the pretrained vision and language model and use the adapters for sign language domain transfer.
This approach ensures our model's primary focus on capturing and utilizing sign language features and then leverages the language model's linguistic ability to adapt the features to spoken language translation. The result is an architecture tailored to SLT, enabling effective translation from video input to textual output.
In our framework, the spatial model and decoder are pretrained from large-scale datasets while the sign encoder is randomly initialized. We therefore develop a gloss-free strategy to pretrain the sign encoder.

\subsection{Pretraining Stage}

\paragraph{Pseudo-gloss generation.}
\label{sec:extract}
To perform pseudo-gloss generation, we extract pseudo-gloss from each spoken language sentence using the spaCy natural language processing library \citep{Honnibal_spaCy}. In the case of German (Phoenix14T), this involves lemmatization, while for Chinese (CSL-Daily), it encompasses word segmentation. The preprocessing step enables the extraction of the base forms of words.

Following this step, we apply Parts-of-Speech (POS) tagging, retaining only words categorized as ["NOUN", "NUM", "ADV", "PRON", "PROPN", "ADJ", "VERB"]. This filtering process prioritizes words that are most likely to convey meaningful information, ensuring that our extracted pseudo-glosses are semantically relevant, retain most of the sentence context and have potential gloss correspondence. The lemmatized words (in German) or segmented words (in Chinese) that satisfy this filter are considered as pseudo-glosses. It's important to highlight that our pseudo-glosses are in spoken language order, unlike manually annotated glosses which are in sign order. Consequently, conventional approaches such as the CTC loss \citep{graves2006connectionist} are not suitable for our pseudo-glosses.

\begin{figure*}[ht]
    \centering
    \includegraphics[width=1.0\textwidth]{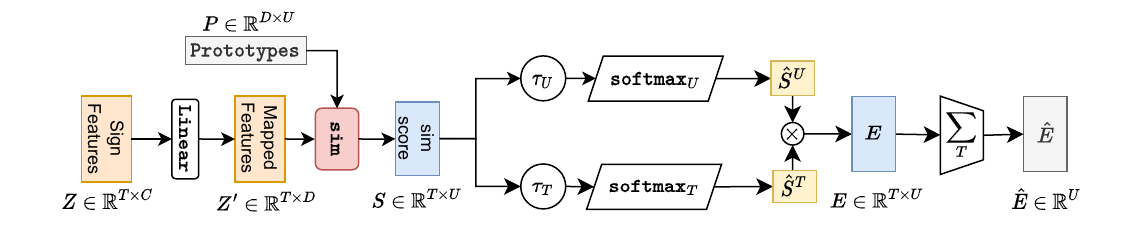}
    \caption{Overview of pretraining process, which takes the sign features as input and predicts the existence of pseudo-glosses.}
    \label{fig:pretraining}
\vspace{-1.0em}
\end{figure*}

\paragraph{Pseudo-gloss pretraining.}
\label{pretraining}
Our goal is to enable the sign encoder to learn visual-linguistic representations. CLIP's ability to unify visual and linguistic domains \citep{radford2021learning} motivates the design of our sign encoder to develop sign features using pseudo-glosses for Sign Language Translation (SLT), as illustrated in Figure \ref{fig:pretraining}.

We generate prototypes for each pseudo-gloss. The aim is to ensure that the sign encoder generates representations that closely align with these prototypes when they are present within the sign video.
These prototypes are initialized with word embeddings obtained from fastText \citep{joulin2016fasttext}, each of dimension $D=300$. Additionally, we include an extra prototype initialized with zeros, which serves as the prototype for sign transitions or non-sign-related components. As a result, the prototype matrix takes the form of $P\in\mathbb{R}^{D\times U}$.

The sign features $Z=\{z_0,z_1,...,z_i,...,z_T\}$ are learned to be aligned with the prototypes if the associated pseudo-gloss exists within the sign video. We therefore project $Z$ to $Z'\in \mathbb{R}^{T\times D}$ through a linear layer and compute the cosine similarity between the prototypes and each of the projected sign features such that:
\begin{equation}
    \label{eq:cosine}
        s_i = \texttt{sim}(z_i', P)= \frac{z_i'\cdot P}{\|\hat{z_i}\| \|P\|}
\end{equation}
which is the cosine similarity score for the $i^{th}$ feature index in $T$ and $S=\{s_0, s_1, ..., s_T\} \in \mathbb{R}^{T\times U}$, $-1\le s_i\le 1$. High scores indicate similarity, and low scores indicate dissimilarity.
We then introduce temporal probability ($\hat{S}^T$) and prototype probability ($\hat{S}^U$) scores using temperature-scaled softmax operations across the time and prototype axis:
\begin{align}
    \label{eq:cosine}
        \hat{S}^T &= \texttt{softmax}_T(S/\tau_T) \\
        \hat{S}^U &= \texttt{softmax}_U(S/\tau_U)
\end{align}
These scores emphasize temporal and class-related aspects of similarity and ensure they fall within the range of 0 to 1. Learnable scaling factors, $\tau_T$ and $\tau_U$, modulate the extent of temporal and class similarity, allowing control of their influence on prototype creation. 
Element-wise multiplication of $\hat{S}^T$ and $\hat{S}^U$ yields $E\in\mathbb{R}^{T\times U}$, forming the basis for prototype localization.
To discern the presence or absence of prototypes, we aggregate $E$ values over the temporal dimension, resulting in $\hat{E}\in\mathbb{R}^{U}$ such that:
\begin{equation}
    \hat{E_{j}} = \sum^T_{i=0}E_{i,j} = \sum^T_{i=0} (\hat{S}_{i,j}^T \times \hat{S}_{i,j}^U)
\end{equation}
 where $0\le E_{i,j} \le 1$.
High $\hat{E}_j$ value for the $j^{th}$ prototype indicates the presence of the prototype, while low values signify the absence.
We employ binary cross-entropy loss to train the model, optimizing the presence or absence of prototypes.
For our case each prototype is assigned as 1 if the pseudo-gloss exists within the sign video and 0 if not.

Note that during pretraining the learned sign representations are temporally invariant, therefore we also add sinusoidal positional encoding before $\texttt{FC}_m$ for the translation task.
The resulting pretraining allows the weights to be initialized with vision priors (spatial model), sign priors (sign encoder) and linguistic priors (decoder). This enables us to then use these models for the downstream translation task.

\section{Results}

\subsection{Datasets and Evaluation Protocol}

We evaluate our approach using two distinct sign language translation datasets.
\textbf{RWTH-PHOENIX-WEATHER-2014T (Phoenix14T)} \citep{camgoz2018neural} is a German Sign Language dataset for sign to spoken language translation tasks. It encompasses translations from German Sign Language to the German language, derived from German weather broadcast videos.
\textbf{CSL-Daily} \citep{zhou2021improving} is a translation dataset focusing on Chinese Sign Language to Chinese. It comprises of lab recorded videos that cover various daily interaction topics, including travel, family life, bank service and shopping.

We evaluate the translation performance using standard metrics commonly employed in sign language translation. These metrics include BLEU (Bilingual Evaluation Understudy) \citep{papineni2002bleu} and ROUGE-L \citep{lin2004rouge} scores. 
BLEU measures the similarity between machine-generated and reference translations based on n-gram overlap. ROUGE-L assesses translation quality by calculating the length of the longest common sub-sequence between generated and reference texts.

\subsection{Training Settings}
\label{sec:trainingsettings}
The model is trained end-to-end with a batch size of 8 on two A100 GPUs, subsampling every second frame. 
Due to memory constraints we apply spatial adapters to the top 3 layers of the spatial model. The sign encoder is a 4 layer transformer with hidden dimension of 512, 8 attention heads and intermediate size of 2048. The temporal downsampling is applied after the 2nd layer. Training is conducted in bfloat16 and Flash attention v2 \citep{dao2023flashattention2} is used to optimise memory usage. We explain further details of the libraries and models in Appendix \ref{library}.
We employ the Adam optimizer \citep{kingma2014adam} with a learning rate of $3\times 10^{-4}$ and weight decay of 0.001. Training spans 100 epochs with gradient clipping of 1.0 and includes a one-cycle cosine learning rate scheduler \citep{loshchilov2016sgdr} with warmup for the initial 5 epochs. Data augmentation techniques, such as color jitter, random resized cropping from $256\times256$ to $224\times224$ pixels, frame rotations and horizontal flips are consistently applied across all frames within video sequences. During evaluation, we use center cropping to $224\times224$ pixels. 

\paragraph{Pretraining.} 
We initialize the prototype ($\tau_U$) and time temperature ($\tau_T$) to 0.1.
In Table~\ref{tab:both}(a), we present the counts of extracted pseudo-glosses for each dataset using the approach described in Section \ref{sec:extract}. For the CSL-Daily dataset, the number of pseudo-glosses significantly exceeds the vocabulary size. This discrepancy arises because the vocabulary is based on Chinese characters, while pseudo-glosses represent word segmentation that convey meaning.
\paragraph{Downstream Translation.} We utilize cross-entropy loss with label smoothing set to 0.1 during training. The LoRA rank and alpha values are both set to 4. During inference, we employ a beam search with a width of 4. 
In Table~\ref{tab:both}(b), we present detailed information regarding the number of trainable parameters for each component of our network during the downstream training phase. Notably, the sign encoder contains a significantly larger number of trainable parameters, emphasizing its role in learning sign features. For the CSL-Daily dataset, we implement additional tokenization processing to handle unknown tokens, as elaborated in Appendix \ref{apx:newtokens}.

\begin{table}[ht]
\begin{center}
    \caption{Training setting with (a) the number of pseudo-glosses during pretraining and (b) parameter counts during downstream translation.}
    \label{tab:both}
    \begin{subtable}{0.5\linewidth}
        \begin{center}
        \label{tab:pseudoglosses}
            \begin{tabular}{lcc}
            \toprule
            \textbf{Dataset} & \textbf{\# Vocab} & \textbf{\# p-glosses} \\
            \midrule
            Phoenix14T & $2,887$ & $2,533$ \\
            CSL-Daily & $2,343$ & $7,918$ \\
            \bottomrule
            \end{tabular}
        \caption{Vocabulary versus pseudo-glosses per dataset}
        \end{center}
    \end{subtable}%
    \begin{subtable}{0.5\linewidth}
        \begin{center}
        \label{tab:trainable}
            \begin{tabular}{lll}
                \toprule
                \textbf{Component} & \textbf{\# Params} & \textbf{\# Trainable} \\
                \midrule
                Spatial         & 22,328,448 & 271,872 \\
                Sign Encoder    & 12,613,632 & 12,613,632 \\
                Decoder     & 1,736,710,528 & 3,803,520 \\
                \midrule
                \textbf{Total}  & 1,771,652,608 & 16,689,024\\
                \bottomrule
            \end{tabular}
        \caption{Number of model parameters during translation}
        \end{center}
    \end{subtable}
\end{center}
\end{table}

\subsection{Comparisons with State-of-the-art Methods}

\paragraph{Results on Phoenix14T.}
In Table~\ref{tab:phx_translation}, we present a comparative analysis between our approach and state-of-the-art methods for sign language translation on Phoenix14T. We observe an approximate improvement of 1.1 BLEU4 on the test set when doing pseudo-gloss pretraining (\textit{PGP}) followed by downstream translation (\textit{Sign2GPT(w/PGP)}). Our gloss-free approach demonstrates substantial improvements in BLEU-{1,2,3} scores, surpassing the performance of previous gloss-free methods and reduces the performance gap between gloss-free and gloss-based SLT. This progress can be attributed to our approach's capability to learn representations for individual pseudo-gloss, in contrast to prior gloss-free methods that primarily focused on sentence-based representations. Notably, even without pretraining (\textit{Sign2GPT}), our approach remains competitive with previous state-of-the-art (SOTA) gloss-free results. We also find that downstream training with only the trainable decoder parameters and extract sign features from our frozen pretrained stage model, \textit{Sign($Z$)2GPT(w/PGP)}, is able to achieve competitive results with under 4 million parameters showing the effectiveness of our learnt sign representations.

\begin{table}[ht]
\centering
\caption{Comparison of test set results on Phoenix14T. We present our gloss-free results for three experimental settings: (1) Without pseudo-gloss pretraining (\textbf{Sign2GPT}), (2) with pseudo-gloss pretraining (\textbf{Sign2GPT(w/PGP)}), and (3) extracted features $Z$ from the frozen spatial and sign encoder model that has been trained with pseudo-gloss pretraining (\textbf{Sign($Z$)2GPT(w/PGP)}).}
\label{tab:phx_translation}
\begin{tabular}{lccccc}
\toprule
\multirow{2}{*}{\textbf{Method}} & \multicolumn{5}{c}{\textbf{Test Set}} \\
\cmidrule(lr){2-6}
 & \textbf{BLEU1} & \textbf{BLEU2} & \textbf{BLEU3} & \textbf{BLEU4} & \textbf{ROUGE}  \\
  \midrule
  \rowcolor{gray!40} \multicolumn{6}{c}{\textbf{Gloss-based}}\\
  \midrule
SL-Transformer \citep{camgoz2020sign}   & $46.61$ & $33.73$ & $26.19$ & $21.32$ & $-$\\
BN-TIN-Transf.+BT  \citep{zhou2021improving}   &  $50.80$ & $37.75$ & $29.72$ & $24.32$ & $49.54$\\
MMTLB \citep{chen2022simple}   & $53.97$ & $41.75$ & $33.84$ & $28.39$ & $52.65$\\
SLTU$_{\texttt{NET}}$  \citep{zhang2023sltunet}  & $52.92$ & $41.76$ & $33.99$ & $28.47$ &  $52.11$\\
TwoStream-SLT \citep{chen2022two}   & $54.90$ & $42.43$ & $34.46$ & $28.95$ & $53.48$\\

 \midrule
 \rowcolor{gray!40}
 \multicolumn{6}{c}{\textbf{Gloss-free}}\\
\midrule
NSLT \citep{camgoz2018neural}   & $29.86$ & $17.52$ & $11.96$ & $9.00$ & $30.70$ \\
TSPNet \citep{li2020tspnet}     & $36.10$ & $23.12$ & $16.88$ & $13.41$ & $34.96$ \\
CSGCR \citep{zhao2021conditional}        & $36.71$ & $25.40$ & $18.86$ & $15.18$ & $38.85$ \\
GASLT \citep{yin2023gloss}         & $39.07$ & $26.74$ & $21.86$ & $15.74$ & $39.86$ \\
GFSLT \citep{zhou2023gloss}   & $41.39$ & $31.00$ & $24.20$ & $19.66$ & $40.93$ \\
GFSLT-VLP \citep{zhou2023gloss}   & $43.71$ & $33.18$ & $26.11$ & $21.44$ & $42.49$ \\
\midrule
\textbf{Sign2GPT}    & $45.43$ & $32.03$ & $24.23$ & $19.42$ & $45.23$ \\
\textbf{Sign2GPT(w/PGP)}  & \textbf{49.54} & \textbf{35.96} & \textbf{28.83} & \textbf{22.52} & \textbf{48.90} \\
\midrule
\textbf{Sign($Z$)2GPT(w/PGP)}& $47.06$ & $33.61$ & $25.85$ & $20.93$ & $47.11$ \\
\bottomrule
\end{tabular}
\end{table}

\paragraph{Results on CSL-Daily.}
In Table~\ref{tab:csl_translation}, we present the results of our SLT experiments on the CSL-Daily dataset. 
We observe a substantial performance increase in BLEU-4 compared to previous SOTA gloss-free results, with an approximate improvement of 4.4 BLEU-4 when utilizing our pretraining (\textit{Sign2GPT(w/PGP)}). \textit{Sign2GPT(w/PGP)} significantly outperforms training without PGP (\textit{Sign2GPT}), with a notable 3.5 BLEU-4 improvement on the CSL-Daily dataset.

\begin{table}[ht]
\centering
\caption{Comparison of test set results on the CSL-Daily. We present gloss-free results for three experimental settings: (1) Without pseudo-gloss pretraining (\textbf{Sign2GPT}), (2) with pseudo-gloss pretraining (\textbf{Sign2GPT(w/PGP)}), and (3) extracted features $Z$ from the frozen spatial and sign encoder model that has been trained with pseudo-gloss pretraining (\textbf{Sign($Z$)2GPT(w/PGP)}).}
\label{tab:csl_translation}
\begin{tabular}{lccccc}
\toprule
\multirow{2}{*}{\textbf{Method}} & \multicolumn{5}{c}{\textbf{Test Set}} \\
\cmidrule(lr){2-6}
 & \textbf{BLEU1} & \textbf{BLEU2} & \textbf{BLEU3} & \textbf{BLEU4} & \textbf{ROUGE}  \\
 \midrule
  \rowcolor{gray!40} \multicolumn{6}{c}{\textbf{Gloss-based}}\\
  \midrule
SL-Transformer \citep{camgoz2020sign}   & $37.38$ & $24.36$ & $16.55$ & $11.79$ & $36.74$\\
BN-TIN-Transf.+BT  \citep{zhou2021improving}   &  $51.42$ & $37.26$ & $27.76$ & $21.34$ & $49.31$\\
MMTLB \citep{chen2022simple}   & $53.31$ & $40.41$ & $30.87$ & $23.92$ & $53.25$\\
SLTU$_{\texttt{NET}}$  \citep{zhang2023sltunet}  & $54.98$ & $41.44$ & $31.84$ & $25.01$& $54.08$\\
TwoStream-SLT \citep{chen2022two}   & $55.44$ & $42.59$ & $32.87$ & $25.79$ & $55.72$\\

 \midrule
 \rowcolor{gray!40}
 \multicolumn{6}{c}{\textbf{Gloss-free}}\\
\midrule
GASLT \citep{yin2023gloss}      & $19.90$ & $9.94$  & $5.98$  & $4.07$  & $20.35$ \\
NSLT \citep{camgoz2018neural}  & $34.16$ & $19.57$ & $11.84$ & $7.56$  & $34.54$ \\
GFSLT \citep{zhou2023gloss}     & $37.69$ & $23.28$ & $14.93$ & $9.88$ & $35.16$ \\
GFSLT-VLP \citep{zhou2023gloss}     & $39.37$ & $24.93$ & $16.26$ & $11.00$ & $36.44$ \\
\midrule
\textbf{Sign2GPT}    & $34.80$ & $24.00$ & $17.27$ & $12.96$ & $41.12$ \\
\textbf{Sign2GPT(w/PGP)} & $\textbf{41.75}$ & $\textbf{28.73}$ & $\textbf{20.60}$ & $\textbf{15.40}$ & $\textbf{42.36}$ \\
\midrule
\textbf{Sign($Z$)2GPT(w/PGP)}  & $32.73$ & $20.52$ & $13.75$ & $9.73$ & $33.39$ \\
\bottomrule
\end{tabular}
\end{table}

\subsection{Qualitative Results}

\begin{figure*}[ht]
    \centering
    \includegraphics[trim={0 0.8em 0 0.5em},clip,width=\textwidth]{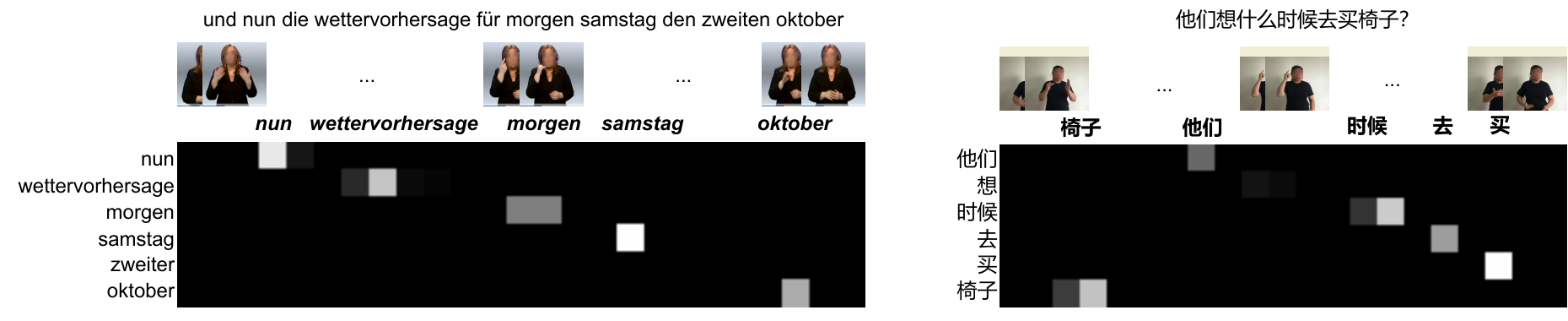}
    \caption{Visualizations of the localization capabilities of our pretraining stage. We visualize only the pseudo-glosses from the target sentence (y-axis) over time (x-axis), with whiter regions indicating a higher probability of the pseudo-gloss occurring during the time segment. We also display the localized gloss (under the video frames) based on a threshold of 0.2 on $E$.}
    \label{fig:viz}
\vspace{-1em}
\end{figure*}

We visually demonstrate the effectiveness of our pseudo-gloss pretraining phase for sign localization in Figure \ref{fig:viz} and Appendix \ref{ap:qual}. To achieve this, we leverage the output values of $E$, with dimensions $T\times U$. The values in this matrix vary between 0 (indicating the absence of a sign) and 1 (indicating the presence of a sign), reflecting the temporal occurrence of pseudo-glosses.
Our observations suggest that our pretraining approach holds promise for sign spotting, even in scenarios where precise localization is not available. This aspect may warrant further exploration as a potential avenue for future research.

\subsection{Ablation Study}
We conduct our ablation studies on the Phoenix14T dataset, evaluating the BLEU-4 score on the development set.
In Table \ref{tab:ablation}, we present the results of architectural modifications made to our network. For our initial study, we conducted experiments without pretraining Sign2GPT. We observed significant performance improvements when incorporating spatial adapter. Conversely, the use of global attention led to a reduction in model performance compared to local attention.
Furthermore, we investigated the impact of downsampling on the model's performance. While downsampling showed only marginal improvements in terms of the BLEU-4 score, it notably reduced computational complexity. This reduction is due to only half the number of features being passed through the decoder during training. 

Our pretraining generates representations for pseudo-glosses, which inherently removes temporal information from the output sign representations. Subsequently, we investigated methods to reintroduce temporal information into the model using various approaches as described in Section \ref{pretraining}.
We explored learnable approaches for positional embeddings with both zero and random initialization. However, these approaches yielded minimal to no discernible benefits. In contrast, the utilization of sinusoidal positional embeddings demonstrated substantial performance improvements.

\begin{table}[ht]
\centering
\caption{Ablation of results on the Phoenix14T dataset showing different architecture changes with no pseudo-gloss pretraining (\textbf{Sign2GPT}) and with pseudo-gloss pretraining (\textbf{Sign2GPT(w/PGP)}).}
\label{tab:ablation}
\begin{tabular}{lc}
    \toprule
     \textbf{Architecture}  &  \textbf{BLEU4} \\
    \midrule
    \textbf{Sign2GPT} &  \\
    Spatial Adapters + Local Attention + Downsampling & $\textbf{19.55}$ \\
    \hspace{0.2 cm} $\cdot$ No Spatial Adapters  & $16.38$  \\
    \hspace{0.2 cm} $\cdot$ No Local Attention (+ Global Attention) & $18.56$  \\
    \hspace{0.2 cm} $\cdot$ No Downsampling on Sign Encoder & $19.30$  \\
    \midrule
    \textbf{Sign2GPT(w/PGP)} &  \\
    \hspace{0.2 cm} $\cdot$ No positional& $21.68$  \\
    \hspace{0.2 cm} $\cdot$ Learnable positional (zero init)  & $21.89$  \\
    \hspace{0.2 cm} $\cdot$ Learnable positional (random init)  & $21.16$  \\
    \hspace{0.2 cm} $\cdot$ Sinusoidal positional & $\textbf{23.20}$  \\
    \bottomrule
\end{tabular}
\vspace{-1em}
\end{table}

\section{Conclusions}

In this paper, we have presented a novel approach to address the challenging problem of Sign Translation in a gloss-free setting. 
Our method, Sign2GPT, demonstrates significant performance improvements over existing state-of-the-art techniques on the Phoenix14T and CSL-Daily datasets. 
We introduce a novel pretraining strategy that learns from pseudo-glosses which are generated automatically to learn word-level sign features, thereby allowing our sign encoder to be effectively pretrained without the use of manually annotated glosses. 
Moreover, the proposed Sign2GPT architecture presents a promising direction for the exploration of fusing visual features to spoken language models for sign language recognition and translation tasks.

\section*{Reproducibility Statement}

To facilitate reproducibility, we have provided details of the training settings in Section \ref{sec:trainingsettings} with additional details of the libraries we used for the pretrained models in Appendix \ref{library}. We also give further details of the CSL tokenization and post processing to address issues with unknown tokens in Appendix \ref{apx:newtokens}.

\section*{Acknowledgements}
This work was supported by the EPSRC project ExTOL (EP/R03298X/1), SNSF project ’SMILE II’ (CRSII5 193686), European Union’s Horizon2020 programme (’EASIER’ grant agreement 101016982) and the Innosuisse IICT Flagship (PFFS-21-47). This work reflects only the authors view and the Commission is not responsible for any use that may be made of the information it contains. Neither Necati Cihan Camgoz nor Meta were involved in the model training, evaluation, or use of the datasets.

\bibliography{iclr2024_conference}

\begin{thebibliography}{58}
\providecommand{\natexlab}[1]{#1}
\providecommand{\url}[1]{\texttt{#1}}
\expandafter\ifx\csname urlstyle\endcsname\relax
  \providecommand{\doi}[1]{doi: #1}\else
  \providecommand{\doi}{doi: \begingroup \urlstyle{rm}\Url}\fi

\bibitem[Alayrac et~al.(2022)Alayrac, Donahue, Luc, Miech, Barr, Hasson, Lenc, Mensch, Millican, Reynolds, et~al.]{alayrac2022flamingo}
Jean-Baptiste Alayrac, Jeff Donahue, Pauline Luc, Antoine Miech, Iain Barr, Yana Hasson, Karel Lenc, Arthur Mensch, Katherine Millican, Malcolm Reynolds, et~al.
\newblock Flamingo: a visual language model for few-shot learning.
\newblock \emph{Advances in Neural Information Processing Systems}, 35:\penalty0 23716--23736, 2022.

\bibitem[Albanie et~al.(2020)Albanie, Varol, Momeni, Afouras, Chung, Fox, and Zisserman]{albanie2020bsl}
Samuel Albanie, G{\"u}l Varol, Liliane Momeni, Triantafyllos Afouras, Joon~Son Chung, Neil Fox, and Andrew Zisserman.
\newblock Bsl-1k: Scaling up co-articulated sign language recognition using mouthing cues.
\newblock In \emph{Computer Vision--ECCV 2020: 16th European Conference, Glasgow, UK, August 23--28, 2020, Proceedings, Part XI 16}, pp.\  35--53. Springer, 2020.

\bibitem[Braem \& Sutton-Spence(2001)Braem and Sutton-Spence]{braem2001hands}
P~Boyes Braem and RL~Sutton-Spence.
\newblock \emph{The Hands Are The Head of The Mouth. The Mouth as Articulator in Sign Languages}.
\newblock Hamburg: Signum Press, 2001.

\bibitem[Camgoz et~al.(2018)Camgoz, Hadfield, Koller, Ney, and Bowden]{camgoz2018neural}
Necati~Cihan Camgoz, Simon Hadfield, Oscar Koller, Hermann Ney, and Richard Bowden.
\newblock Neural sign language translation.
\newblock In \emph{Proceedings of the IEEE conference on computer vision and pattern recognition}, pp.\  7784--7793, 2018.

\bibitem[Camgoz et~al.(2020{\natexlab{a}})Camgoz, Koller, Hadfield, and Bowden]{camgoz2020multi}
Necati~Cihan Camgoz, Oscar Koller, Simon Hadfield, and Richard Bowden.
\newblock Multi-channel transformers for multi-articulatory sign language translation.
\newblock In \emph{Computer Vision--ECCV 2020 Workshops: Glasgow, UK, August 23--28, 2020, Proceedings, Part IV 16}, pp.\  301--319. Springer, 2020{\natexlab{a}}.

\bibitem[Camgoz et~al.(2020{\natexlab{b}})Camgoz, Koller, Hadfield, and Bowden]{camgoz2020sign}
Necati~Cihan Camgoz, Oscar Koller, Simon Hadfield, and Richard Bowden.
\newblock Sign language transformers: Joint end-to-end sign language recognition and translation.
\newblock In \emph{Proceedings of the IEEE/CVF conference on computer vision and pattern recognition}, pp.\  10023--10033, 2020{\natexlab{b}}.

\bibitem[Carreira \& Zisserman(2017)Carreira and Zisserman]{carreira2017quo}
Joao Carreira and Andrew Zisserman.
\newblock Quo vadis, action recognition? a new model and the kinetics dataset.
\newblock In \emph{proceedings of the IEEE Conference on Computer Vision and Pattern Recognition}, pp.\  6299--6308, 2017.

\bibitem[Chen et~al.(2022{\natexlab{a}})Chen, Wei, Sun, Wu, and Lin]{chen2022simple}
Yutong Chen, Fangyun Wei, Xiao Sun, Zhirong Wu, and Stephen Lin.
\newblock A simple multi-modality transfer learning baseline for sign language translation.
\newblock In \emph{Proceedings of the IEEE/CVF Conference on Computer Vision and Pattern Recognition}, pp.\  5120--5130, 2022{\natexlab{a}}.

\bibitem[Chen et~al.(2022{\natexlab{b}})Chen, Zuo, Wei, Wu, Liu, and Mak]{chen2022two}
Yutong Chen, Ronglai Zuo, Fangyun Wei, Yu~Wu, Shujie Liu, and Brian Mak.
\newblock Two-stream network for sign language recognition and translation.
\newblock \emph{Advances in Neural Information Processing Systems}, 35:\penalty0 17043--17056, 2022{\natexlab{b}}.

\bibitem[Cui et~al.(2017)Cui, Liu, and Zhang]{cui2017recurrent}
Runpeng Cui, Hu~Liu, and Changshui Zhang.
\newblock Recurrent convolutional neural networks for continuous sign language recognition by staged optimization.
\newblock In \emph{Proceedings of the IEEE conference on computer vision and pattern recognition}, pp.\  7361--7369, 2017.

\bibitem[Dao(2023)]{dao2023flashattention2}
Tri Dao.
\newblock Flash{A}ttention-2: Faster attention with better parallelism and work partitioning.
\newblock 2023.

\bibitem[Dosovitskiy et~al.(2020)Dosovitskiy, Beyer, Kolesnikov, Weissenborn, Zhai, Unterthiner, Dehghani, Minderer, Heigold, Gelly, et~al.]{dosovitskiy2020image}
Alexey Dosovitskiy, Lucas Beyer, Alexander Kolesnikov, Dirk Weissenborn, Xiaohua Zhai, Thomas Unterthiner, Mostafa Dehghani, Matthias Minderer, Georg Heigold, Sylvain Gelly, et~al.
\newblock An image is worth 16x16 words: Transformers for image recognition at scale.
\newblock \emph{arXiv preprint arXiv:2010.11929}, 2020.

\bibitem[Gal et~al.(2022)Gal, Alaluf, Atzmon, Patashnik, Bermano, Chechik, and Cohen-Or]{gal2022image}
Rinon Gal, Yuval Alaluf, Yuval Atzmon, Or~Patashnik, Amit~H Bermano, Gal Chechik, and Daniel Cohen-Or.
\newblock An image is worth one word: Personalizing text-to-image generation using textual inversion.
\newblock \emph{arXiv preprint arXiv:2208.01618}, 2022.

\bibitem[Gan et~al.()Gan, Yin, Jiang, Xia, Xie, and Lu]{gancontrastive}
Shiwei Gan, Yafeng Yin, Zhiwei Jiang, Kang Xia, Lei Xie, and Sanglu Lu.
\newblock Contrastive learning for sign language recognition and translation.

\bibitem[Gao et~al.(2023)Gao, Han, Zhang, Lin, Geng, Zhou, Zhang, Lu, He, Yue, et~al.]{gao2023llama}
Peng Gao, Jiaming Han, Renrui Zhang, Ziyi Lin, Shijie Geng, Aojun Zhou, Wei Zhang, Pan Lu, Conghui He, Xiangyu Yue, et~al.
\newblock Llama-adapter v2: Parameter-efficient visual instruction model.
\newblock \emph{arXiv preprint arXiv:2304.15010}, 2023.

\bibitem[Graves et~al.(2006)Graves, Fern{\'a}ndez, Gomez, and Schmidhuber]{graves2006connectionist}
Alex Graves, Santiago Fern{\'a}ndez, Faustino Gomez, and J{\"u}rgen Schmidhuber.
\newblock Connectionist temporal classification: labelling unsegmented sequence data with recurrent neural networks.
\newblock In \emph{Proceedings of the 23rd international conference on Machine learning}, pp.\  369--376, 2006.

\bibitem[Hewitt(2021)]{hewitt2021initializing}
John Hewitt.
\newblock Initializing new word embeddings for pretrained language models.
\newblock \url{https:/nlp.stanford.edu/~johnhew//vocab-expansion.html}, 2021.

\bibitem[Honnibal et~al.(2020)Honnibal, Montani, Van~Landeghem, and Boyd]{Honnibal_spaCy}
Matthew Honnibal, Ines Montani, Sofie Van~Landeghem, and Adriane Boyd.
\newblock {spaCy: Industrial-strength Natural Language Processing in Python}.
\newblock 2020.
\newblock \doi{10.5281/zenodo.1212303}.

\bibitem[Hu et~al.(2021)Hu, Shen, Wallis, Allen-Zhu, Li, Wang, Wang, and Chen]{hu2021lora}
Edward~J Hu, Yelong Shen, Phillip Wallis, Zeyuan Allen-Zhu, Yuanzhi Li, Shean Wang, Lu~Wang, and Weizhu Chen.
\newblock Lora: Low-rank adaptation of large language models.
\newblock \emph{arXiv preprint arXiv:2106.09685}, 2021.

\bibitem[Jiang et~al.(2021)Jiang, Sun, Wang, Bai, Li, and Fu]{jiang2021skeleton}
Songyao Jiang, Bin Sun, Lichen Wang, Yue Bai, Kunpeng Li, and Yun Fu.
\newblock Skeleton aware multi-modal sign language recognition.
\newblock In \emph{Proceedings of the IEEE/CVF conference on computer vision and pattern recognition}, pp.\  3413--3423, 2021.

\bibitem[Joulin et~al.(2016)Joulin, Grave, Bojanowski, Douze, J{\'e}gou, and Mikolov]{joulin2016fasttext}
Armand Joulin, Edouard Grave, Piotr Bojanowski, Matthijs Douze, H{\'e}rve J{\'e}gou, and Tomas Mikolov.
\newblock Fasttext.zip: Compressing text classification models.
\newblock \emph{arXiv preprint arXiv:1612.03651}, 2016.

\bibitem[Joze \& Koller(2018)Joze and Koller]{joze2018ms}
Hamid Reza~Vaezi Joze and Oscar Koller.
\newblock Ms-asl: A large-scale data set and benchmark for understanding american sign language.
\newblock \emph{arXiv preprint arXiv:1812.01053}, 2018.

\bibitem[Kingma \& Ba(2014)Kingma and Ba]{kingma2014adam}
Diederik~P Kingma and Jimmy Ba.
\newblock Adam: A method for stochastic optimization.
\newblock \emph{arXiv preprint arXiv:1412.6980}, 2014.

\bibitem[Koller et~al.(2015)Koller, Forster, and Ney]{koller2015continuous}
Oscar Koller, Jens Forster, and Hermann Ney.
\newblock Continuous sign language recognition: Towards large vocabulary statistical recognition systems handling multiple signers.
\newblock \emph{Computer Vision and Image Understanding}, 141:\penalty0 108--125, 2015.

\bibitem[Li et~al.(2020{\natexlab{a}})Li, Rodriguez, Yu, and Li]{li2020word}
Dongxu Li, Cristian Rodriguez, Xin Yu, and Hongdong Li.
\newblock Word-level deep sign language recognition from video: A new large-scale dataset and methods comparison.
\newblock In \emph{Proceedings of the IEEE/CVF winter conference on applications of computer vision}, pp.\  1459--1469, 2020{\natexlab{a}}.

\bibitem[Li et~al.(2020{\natexlab{b}})Li, Xu, Yu, Zhang, Swift, Suominen, and Li]{li2020tspnet}
Dongxu Li, Chenchen Xu, Xin Yu, Kaihao Zhang, Benjamin Swift, Hanna Suominen, and Hongdong Li.
\newblock Tspnet: Hierarchical feature learning via temporal semantic pyramid for sign language translation.
\newblock \emph{Advances in Neural Information Processing Systems}, 33:\penalty0 12034--12045, 2020{\natexlab{b}}.

\bibitem[Lin(2004)]{lin2004rouge}
Chin-Yew Lin.
\newblock Rouge: A package for automatic evaluation of summaries.
\newblock In \emph{Text summarization branches out}, pp.\  74--81, 2004.

\bibitem[Lin et~al.(2021)Lin, Mihaylov, Artetxe, Wang, Chen, Simig, Ott, Goyal, Bhosale, Du, et~al.]{lin2021few}
Xi~Victoria Lin, Todor Mihaylov, Mikel Artetxe, Tianlu Wang, Shuohui Chen, Daniel Simig, Myle Ott, Naman Goyal, Shruti Bhosale, Jingfei Du, et~al.
\newblock Few-shot learning with multilingual language models.
\newblock \emph{arXiv preprint arXiv:2112.10668}, 2021.

\bibitem[Liu et~al.(2020)Liu, Gu, Goyal, Li, Edunov, Ghazvininejad, Lewis, and Zettlemoyer]{liu2020multilingual}
Yinhan Liu, Jiatao Gu, Naman Goyal, Xian Li, Sergey Edunov, Marjan Ghazvininejad, Mike Lewis, and Luke Zettlemoyer.
\newblock Multilingual denoising pre-training for neural machine translation.
\newblock \emph{Transactions of the Association for Computational Linguistics}, 8:\penalty0 726--742, 2020.

\bibitem[Loshchilov \& Hutter(2016)Loshchilov and Hutter]{loshchilov2016sgdr}
Ilya Loshchilov and Frank Hutter.
\newblock Sgdr: Stochastic gradient descent with warm restarts.
\newblock \emph{arXiv preprint arXiv:1608.03983}, 2016.

\bibitem[Momeni et~al.(2022)Momeni, Bull, Prajwal, Albanie, Varol, and Zisserman]{momeni2022automatic}
Liliane Momeni, Hannah Bull, KR~Prajwal, Samuel Albanie, G{\"u}l Varol, and Andrew Zisserman.
\newblock Automatic dense annotation of large-vocabulary sign language videos.
\newblock In \emph{European Conference on Computer Vision}, pp.\  671--690. Springer, 2022.

\bibitem[Oquab et~al.(2023)Oquab, Darcet, Moutakanni, Vo, Szafraniec, Khalidov, Fernandez, Haziza, Massa, El-Nouby, et~al.]{oquab2023dinov2}
Maxime Oquab, Timoth{\'e}e Darcet, Th{\'e}o Moutakanni, Huy Vo, Marc Szafraniec, Vasil Khalidov, Pierre Fernandez, Daniel Haziza, Francisco Massa, Alaaeldin El-Nouby, et~al.
\newblock Dinov2: Learning robust visual features without supervision.
\newblock \emph{arXiv preprint arXiv:2304.07193}, 2023.

\bibitem[Papineni et~al.(2002)Papineni, Roukos, Ward, and Zhu]{papineni2002bleu}
Kishore Papineni, Salim Roukos, Todd Ward, and Wei-Jing Zhu.
\newblock Bleu: a method for automatic evaluation of machine translation.
\newblock In \emph{Proceedings of the 40th annual meeting of the Association for Computational Linguistics}, pp.\  311--318, 2002.

\bibitem[Radford et~al.(2018)Radford, Narasimhan, Salimans, Sutskever, et~al.]{radford2018improving}
Alec Radford, Karthik Narasimhan, Tim Salimans, Ilya Sutskever, et~al.
\newblock Improving language understanding by generative pre-training.
\newblock 2018.

\bibitem[Radford et~al.(2021)Radford, Kim, Hallacy, Ramesh, Goh, Agarwal, Sastry, Askell, Mishkin, Clark, et~al.]{radford2021learning}
Alec Radford, Jong~Wook Kim, Chris Hallacy, Aditya Ramesh, Gabriel Goh, Sandhini Agarwal, Girish Sastry, Amanda Askell, Pamela Mishkin, Jack Clark, et~al.
\newblock Learning transferable visual models from natural language supervision.
\newblock In \emph{International conference on machine learning}, pp.\  8748--8763. PMLR, 2021.

\bibitem[Raffel et~al.(2020)Raffel, Shazeer, Roberts, Lee, Narang, Matena, Zhou, Li, and Liu]{raffel2020exploring}
Colin Raffel, Noam Shazeer, Adam Roberts, Katherine Lee, Sharan Narang, Michael Matena, Yanqi Zhou, Wei Li, and Peter~J Liu.
\newblock Exploring the limits of transfer learning with a unified text-to-text transformer.
\newblock \emph{The Journal of Machine Learning Research}, 21\penalty0 (1):\penalty0 5485--5551, 2020.

\bibitem[Roich et~al.(2022)Roich, Mokady, Bermano, and Cohen-Or]{roich2022pivotal}
Daniel Roich, Ron Mokady, Amit~H Bermano, and Daniel Cohen-Or.
\newblock Pivotal tuning for latent-based editing of real images.
\newblock \emph{ACM Transactions on Graphics (TOG)}, 42\penalty0 (1):\penalty0 1--13, 2022.

\bibitem[Ruiz et~al.(2022)Ruiz, Li, Jampani, Pritch, Rubinstein, and Aberman]{ruiz2022dreambooth}
Nataniel Ruiz, Yuanzhen Li, Varun Jampani, Yael Pritch, Michael Rubinstein, and Kfir Aberman.
\newblock Dreambooth: Fine tuning text-to-image diffusion models for subject-driven generation.
\newblock \emph{arXiv preprint arXiv:2208.12242}, 2022.

\bibitem[Shi et~al.(2022)Shi, Brentari, Shakhnarovich, and Livescu]{shi2022open}
Bowen Shi, Diane Brentari, Greg Shakhnarovich, and Karen Livescu.
\newblock Open-domain sign language translation learned from online video.
\newblock \emph{arXiv preprint arXiv:2205.12870}, 2022.

\bibitem[Sincan \& Keles(2020)Sincan and Keles]{sincan2020autsl}
Ozge~Mercanoglu Sincan and Hacer~Yalim Keles.
\newblock Autsl: A large scale multi-modal turkish sign language dataset and baseline methods.
\newblock \emph{IEEE Access}, 8:\penalty0 181340--181355, 2020.

\bibitem[Sincan et~al.(2023)Sincan, Camgoz, and Bowden]{sincan2023context}
Ozge~Mercanoglu Sincan, Necati~Cihan Camgoz, and Richard Bowden.
\newblock Is context all you need? scaling neural sign language translation to large domains of discourse.
\newblock In \emph{Proceedings of the IEEE/CVF International Conference on Computer Vision}, pp.\  1955--1965, 2023.

\bibitem[Soomro et~al.(2012)Soomro, Zamir, and Shah]{soomro2012ucf101}
Khurram Soomro, Amir~Roshan Zamir, and Mubarak Shah.
\newblock Ucf101: A dataset of 101 human actions classes from videos in the wild.
\newblock \emph{arXiv preprint arXiv:1212.0402}, 2012.

\bibitem[Tran et~al.(2018)Tran, Wang, Torresani, Ray, LeCun, and Paluri]{tran2018closer}
Du~Tran, Heng Wang, Lorenzo Torresani, Jamie Ray, Yann LeCun, and Manohar Paluri.
\newblock A closer look at spatiotemporal convolutions for action recognition.
\newblock In \emph{Proceedings of the IEEE conference on Computer Vision and Pattern Recognition}, pp.\  6450--6459, 2018.

\bibitem[Uthus et~al.(2023)Uthus, Tanzer, and Georg]{uthus2023youtube}
David Uthus, Garrett Tanzer, and Manfred Georg.
\newblock Youtube-asl: A large-scale, open-domain american sign language-english parallel corpus.
\newblock \emph{arXiv preprint arXiv:2306.15162}, 2023.

\bibitem[Vaswani et~al.(2017)Vaswani, Shazeer, Parmar, Uszkoreit, Jones, Gomez, Kaiser, and Polosukhin]{vaswani2017attention}
Ashish Vaswani, Noam Shazeer, Niki Parmar, Jakob Uszkoreit, Llion Jones, Aidan~N Gomez, {\L}ukasz Kaiser, and Illia Polosukhin.
\newblock Attention is all you need.
\newblock \emph{Advances in neural information processing systems}, 30, 2017.

\bibitem[Wong et~al.(2023)Wong, Camgoz, and Bowden]{wong2023learnt}
Ryan Wong, Necati~Cihan Camgoz, and Richard Bowden.
\newblock Learnt contrastive concept embeddings for sign recognition.
\newblock In \emph{Proceedings of the IEEE/CVF International Conference on Computer Vision}, pp.\  1945--1954, 2023.

\bibitem[Yan et~al.(2018)Yan, Xiong, and Lin]{yan2018spatial}
Sijie Yan, Yuanjun Xiong, and Dahua Lin.
\newblock Spatial temporal graph convolutional networks for skeleton-based action recognition.
\newblock In \emph{Proceedings of the AAAI conference on artificial intelligence}, volume~32, 2018.

\bibitem[Yao et~al.(2023)Yao, Zhou, Feng, Hu, Zhou, and Li]{yao2023sign}
Huijie Yao, Wengang Zhou, Hao Feng, Hezhen Hu, Hao Zhou, and Houqiang Li.
\newblock Sign language translation with iterative prototype.
\newblock \emph{arXiv preprint arXiv:2308.12191}, 2023.

\bibitem[Ye et~al.(2023)Ye, Jiao, Wang, Tu, and Xiong]{ye2023cross}
Jinhui Ye, Wenxiang Jiao, Xing Wang, Zhaopeng Tu, and Hui Xiong.
\newblock Cross-modality data augmentation for end-to-end sign language translation.
\newblock \emph{arXiv preprint arXiv:2305.11096}, 2023.

\bibitem[Yin et~al.(2023)Yin, Zhong, Tang, Jin, Jin, and Zhao]{yin2023gloss}
Aoxiong Yin, Tianyun Zhong, Li~Tang, Weike Jin, Tao Jin, and Zhou Zhao.
\newblock Gloss attention for gloss-free sign language translation.
\newblock In \emph{Proceedings of the IEEE/CVF Conference on Computer Vision and Pattern Recognition}, pp.\  2551--2562, 2023.

\bibitem[Zhang et~al.(2023{\natexlab{a}})Zhang, M{\"u}ller, and Sennrich]{zhang2023sltunet}
Biao Zhang, Mathias M{\"u}ller, and Rico Sennrich.
\newblock Sltunet: A simple unified model for sign language translation.
\newblock \emph{arXiv preprint arXiv:2305.01778}, 2023{\natexlab{a}}.

\bibitem[Zhang et~al.(2023{\natexlab{b}})Zhang, Li, and Bing]{zhang2023video}
Hang Zhang, Xin Li, and Lidong Bing.
\newblock Video-llama: An instruction-tuned audio-visual language model for video understanding.
\newblock \emph{arXiv preprint arXiv:2306.02858}, 2023{\natexlab{b}}.

\bibitem[Zhang et~al.(2023{\natexlab{c}})Zhang, Han, Zhou, Hu, Yan, Lu, Li, Gao, and Qiao]{zhang2023llama}
Renrui Zhang, Jiaming Han, Aojun Zhou, Xiangfei Hu, Shilin Yan, Pan Lu, Hongsheng Li, Peng Gao, and Yu~Qiao.
\newblock Llama-adapter: Efficient fine-tuning of language models with zero-init attention.
\newblock \emph{arXiv preprint arXiv:2303.16199}, 2023{\natexlab{c}}.

\bibitem[Zhao et~al.(2021)Zhao, Qi, Zhou, Duan, Zhou, and Li]{zhao2021conditional}
Jian Zhao, Weizhen Qi, Wengang Zhou, Nan Duan, Ming Zhou, and Houqiang Li.
\newblock Conditional sentence generation and cross-modal reranking for sign language translation.
\newblock \emph{IEEE Transactions on Multimedia}, 24:\penalty0 2662--2672, 2021.

\bibitem[Zheng et~al.(2023)Zheng, Wang, Tan, Li, Wang, Xia, Chen, and Li]{zheng2023cvt}
Jiangbin Zheng, Yile Wang, Cheng Tan, Siyuan Li, Ge~Wang, Jun Xia, Yidong Chen, and Stan~Z Li.
\newblock Cvt-slr: Contrastive visual-textual transformation for sign language recognition with variational alignment.
\newblock In \emph{Proceedings of the IEEE/CVF Conference on Computer Vision and Pattern Recognition}, pp.\  23141--23150, 2023.

\bibitem[Zhou et~al.(2023)Zhou, Chen, Clap{\'e}s, Wan, Liang, Escalera, Lei, and Zhang]{zhou2023gloss}
Benjia Zhou, Zhigang Chen, Albert Clap{\'e}s, Jun Wan, Yanyan Liang, Sergio Escalera, Zhen Lei, and Du~Zhang.
\newblock Gloss-free sign language translation: Improving from visual-language pretraining.
\newblock \emph{arXiv preprint arXiv:2307.14768}, 2023.

\bibitem[Zhou et~al.(2021)Zhou, Zhou, Qi, Pu, and Li]{zhou2021improving}
Hao Zhou, Wengang Zhou, Weizhen Qi, Junfu Pu, and Houqiang Li.
\newblock Improving sign language translation with monolingual data by sign back-translation.
\newblock In \emph{Proceedings of the IEEE/CVF Conference on Computer Vision and Pattern Recognition}, pp.\  1316--1325, 2021.

\bibitem[Zuo et~al.(2023)Zuo, Wei, and Mak]{zuo2023natural}
Ronglai Zuo, Fangyun Wei, and Brian Mak.
\newblock Natural language-assisted sign language recognition.
\newblock In \emph{Proceedings of the IEEE/CVF Conference on Computer Vision and Pattern Recognition}, pp.\  14890--14900, 2023.

\end{thebibliography}
\bibliographystyle{iclr2024_conference}

\appendix
\section{Appendix}

\subsection{Libraries}
\label{library}
\textbf{Spatial Backbone.} The spatial backbone is ViT-S/14 distilled with the pretrained model downloaded from $\texttt{github.com/facebookresearch/dinov2}$.

\textbf{Decoder.} We use the HuggingFace library for the pretrained XGLM transformer model and tokenization, specifically the pretrained weights from $\texttt{facebook/xglm-1.7B}$.

\textbf{Pseudo-gloss generation.} For pseudo-gloss generation we use the SpaCy library. We use the German pipeline loaded from $\texttt{de\_core\_news\_sm}$ for generating our pseudo-glosses for the Phoenix14T dataset. For the CSL-Daily dataset we use $\texttt{zh\_core\_web\_sm}$.

\textbf{FastText embeddings.} We load the fastText embeddings from $\texttt{cc.de.300.bin}$ for the German Phoenix14T dataset and $\texttt{cc.zh.300.bin}$ for the Chinese CSL Dataset.

\subsection{CSL Tokenization and post processing}
\label{apx:newtokens}

By using the pretrained tokenization from XGLM, we encountered unknown tokens during CSL-daily training when Chinese characters were absent from the vocabulary. To address this, we performed vocabulary expansion, utilizing the average embedding technique by \citet{hewitt2021initializing} for initializing new tokens with slight noise. We maintained the consistency of our proposed architecture by keeping these new embeddings frozen throughout training, avoiding the introduction of additional learnable parameters. In total, there are 252 new tokens added to the vocabulary.
We also apply post-processing of replacing punctuation tokens at inference time as the pretrained tokenizer converts punctuation to the Unicode equivalent such that  '?', '!', ':' and ',' are replaced by \chinese{'？', '！', '：'} and \chinese{'，'} respectively.

\subsection{Qualitative Translation Examples}

In Table \ref{tab:phx_translation_text} and \ref{tab:csl_translation_text} we present randomly selected qualitative translation results on Phoenix14T and CSL-Daily respectively. We also display the extracted pseudo-glosses for each of the target sentences which capture the core content of the sentences. Our findings demonstrate that our translation model successfully generates sentences that effectively capture the semantic content of spoken language sentences, albeit with variations in sentence structure.

\begin{table}[ht]
\centering
\caption{Examples of translation results on the Phoenix14T dataset.}
\label{tab:phx_translation_text}
\begin{tabular}{p{0.2\linewidth}|p{0.8\linewidth}}
\toprule
Hypothesis: & im übrigen land scheint häufig die sonne und es gibt nur wenig schauer . \textbf{(in the rest of the country the sun often shines and there are only a few showers .)}\\
Pseudo-glosses: & übrig, gebiet, sonne, nur, locker, wolke \\
Reference: &in den übrigen gebieten viel sonne und nur ein paar lockere wolken . \textbf{(lots of sun in the remaining areas and only a few loose clouds .)} \\
\midrule
Hypothesis: & am tag zwölf grad an der ostsee und bis zu zwanzig grad im süden . (\textbf{twelve degrees a day on the baltic sea and up to twenty degrees in the south .}) \\
Pseudo-glosses: & tag, zwölf, grad, ostsee, zwanzig, grad, niederbayer\\
Reference: & am tag zwölf grad an der ostsee und bis zwanzig grad in niederbayern . (\textbf{on the day twelve degrees on the baltic sea and up to twenty degrees in lower bavaria . })\\
\midrule
Hypothesis: & ich wünsche ihnen noch einen schönen abend und machen sie es gut . \textbf{(i wish you a nice evening and do well .)}\\
Pseudo-glosses: & ich, wünschen, ihnen, schön, abend, machen, sie, es, gut\\
Reference: & ich wünsche ihnen einen schönen abend und machen sie es gut . \textbf{(
i wish you a nice evening and do well .)}\\
\midrule
Hypothesis: & der wind weht schwach bis mäßig aus süd bis südost . \textbf{(
the wind blows weakly to moderately from the south to southeast .)}\\
Pseudo-glosses: & dazu, wehen, schwach, wind, südost, süd\\
Reference: & dazu weht ein schwacher bis mäßiger wind aus südost bis süd . \textbf{(
in addition a weak to moderate wind blows from the southeast to the south .)} \\
\bottomrule
\end{tabular}
\end{table}

\begin{table}[ht]
\centering
\caption{Examples of translation results on the CSL-Daily dataset.}
\label{tab:csl_translation_text}
\begin{tabular}{p{0.2\linewidth}|p{0.8\linewidth}}
\toprule
Hypothesis: & 
\chinese{这个地方不离饭店，走几步就到饭店的门口。}\textbf{(This place is not far from the hotel, just a few steps to the door of the hotel.)}\\
Pseudo-glosses: & \chinese{可以 / 这里 / 不 / 远 / 有 / 饭馆 / 走 / 几 / 分钟 / 就 / 到} \\
Reference: & \chinese{可以，离这里不远有一个饭馆，走几分钟就到了。} \textbf{(Okay, there is a restaurant not far from here, it can be reached in a few minutes' walk.)} \\
\midrule
Hypothesis: & \chinese{公司很远，他为什么不打车呢？} (\textbf{The company is far away, why doesn't he take a taxi?}) \\
Pseudo-glosses: & \chinese{公司 /离家 / 很 / 远 / 他 / 为什么 / 不 / 打车} \\
Reference: & \chinese{公司离家很远，他为什么不打车？} (\textbf{The company is far from home, why doesn't he take a taxi?})\\
\midrule
Hypothesis: & \chinese{我不去爬山，我有事情要去做。} (\textbf{I'm not going to climb mountains, I have things to do.}) \\
Pseudo-glosses: & \chinese{我 / 不 / 去 / 爬山 / 我 / 有事} \\
Reference: & \chinese{我不去爬山，我有事。} (\textbf{
I'm not going to climb the mountain, I have something to do.})\\
\midrule
Hypothesis: & \chinese{我喜欢下雪。} (\textbf{I like snow.}) \\
Pseudo-glosses: & \chinese{我 / 喜欢 / 冬天 / 下雪 / 太 / 美} \\
Reference: & \chinese{我喜欢冬天，下雪太美了。} (\textbf{
I like winter, the snow is so beautiful.})\\
\bottomrule
\end{tabular}
\end{table}

\subsection{Qualitative Pseudo-gloss Localization Examples}
\label{ap:qual}
\begin{figure*}[ht]
    \centering
    \includegraphics[width=0.9\textwidth]{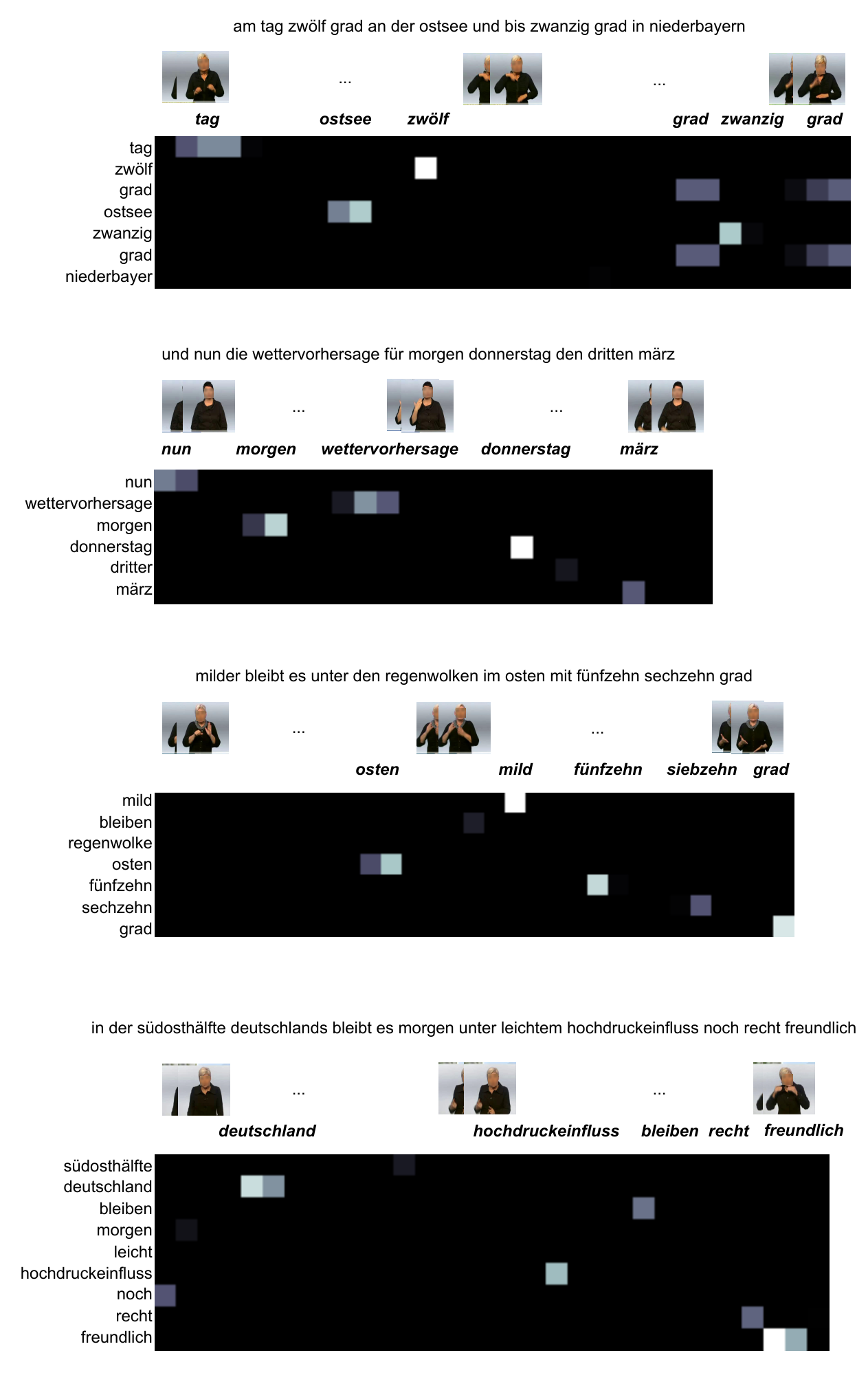}
    \caption{Visualizations of the localization capabilities of our pretraining stage on the pseudo-glosses from the Phoenix14T dataset. We visualize only the pseudo-glosses from the target sentence (y-axis) over time (x-axis), with whiter regions indicating a higher probability of the pseudo-gloss occurring during the time segment. We also display the localized gloss (under the video frames) based on a threshold of 0.2 on $E$.}
    \label{fig:visphx}
\end{figure*}

\begin{figure*}[ht]
    \centering
    \includegraphics[width=0.9\textwidth]{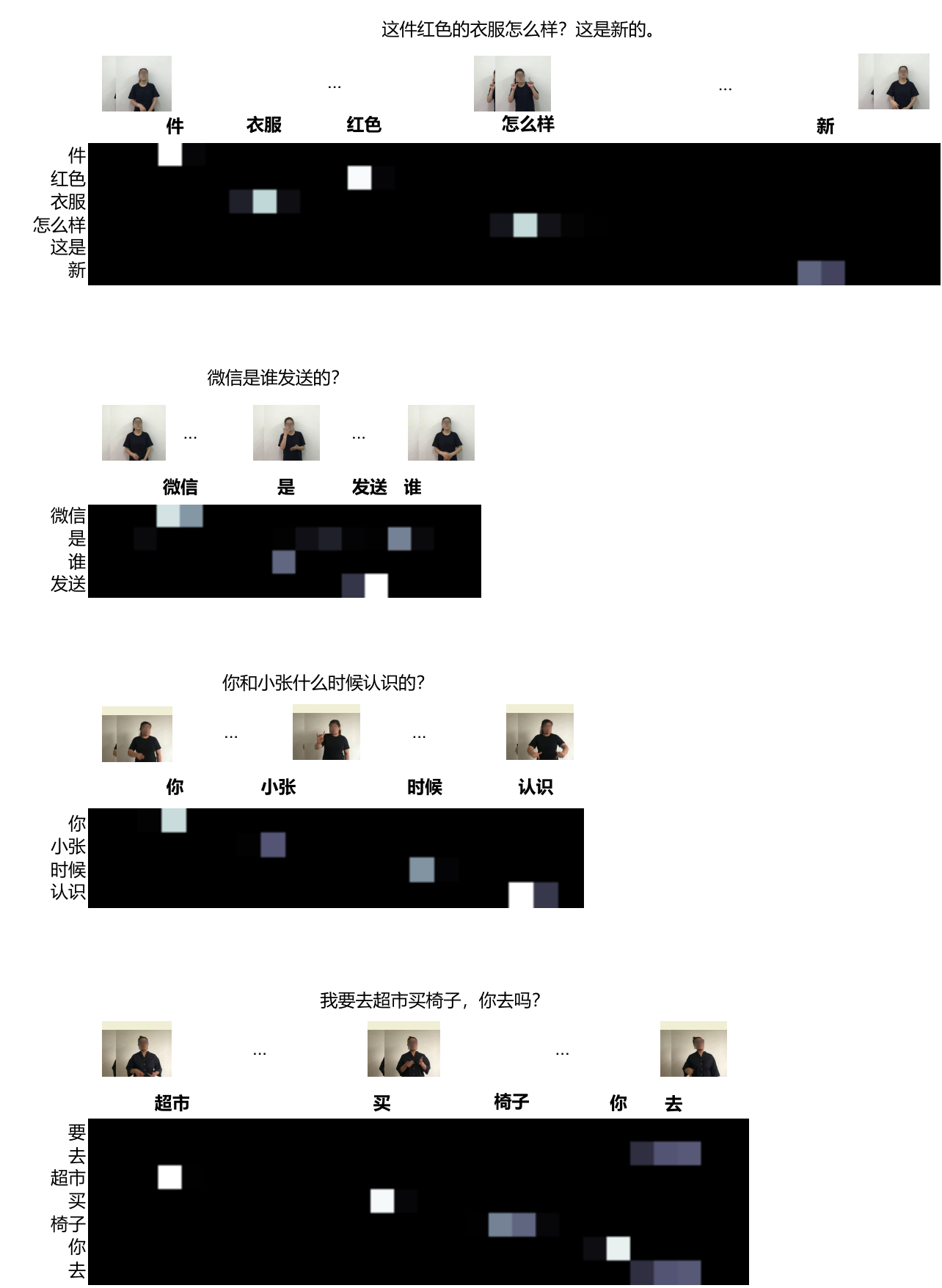}
    \caption{Visualizations of the localization capabilities of our pretraining stage on the pseudo-glosses from the CSL-Daily dataset. We visualize only the pseudo-glosses from the target sentence (y-axis) over time (x-axis), with whiter regions indicating a higher probability of the pseudo-gloss occurring during the time segment. We also display the localized gloss (under the video frames) based on a threshold of 0.2 on $E$.}
    \label{fig:viscsl}
\end{figure*}

We visualize the localization capabilities of our pretraining on Phoenix14T in Figure \ref{fig:visphx} and CSL-Daily in Figure \ref{fig:viscsl}. We notice that our pretraining has automatic localization capabilities irrespective of the order of the pseudo-gloss when using the output $E$. This has the potential future avenues for automatically creating sign-ordered glosses based on thresholds of $E$. We observe that while we provide no information about the localization of the pseudo-glosses, our approach is able to identify the potential glosses and the sign order of these identified glosses.

\section{Additional Ablation Studies}

\subsection{Impact of Spatial Backbone}

In Table \ref{tab:ab_spatial}, we present comparisons between the performance of a ResNet18 spatial backbone and the proposed adapted ViT model within our Sign2GPT architecture. We find that the frozen ViT model trained with DinoV2 pretraining, incorporating adapters, exhibits slightly better performance compared to the ResNet18 backbone.
A noteworthy advantage of the adapted ViT model lies in the application of LoRA instead of fine-tuning all parameters. Consequently, the proposed backbone has under three hundred thousand trainable parameters, a significant reduction compared to the 11 million trainable parameters of the ResNet18 model.

\begin{table}[ht]
\centering
\caption{Ablation of different spatial backbones on CSL-Daily using our Sign2GPT architecture without Pseudo-Gloss Pretraining.}
\label{tab:ab_spatial}
\begin{tabular}{lcc}
\toprule
\multirow{2}{*}{\textbf{Spatial Backbone}} & \multicolumn{2}{c}{\textbf{Test Set}} \\
\cmidrule(lr){2-3}
&  \textbf{BLEU4} & \textbf{ROUGE}  \\
\midrule
ResNet18  & 12.28 & 38.31 \\
DinoV2 (ViT-S/14)      & \textbf{12.96}  & \textbf{41.12} \\
\bottomrule
\end{tabular}
\end{table}

\subsection{Pretraining Performance}

In Table \ref{tab:ab_perf_pretraining}, we demonstrate the performance of our pseudo-gloss pretraining by measuring the precision, recall and F1-score for Phoenix14T and CSL-Daily using a threshold of 0.2.

\begin{table}[ht]
\centering
\caption{Quantitative results of pseudo-gloss pretraining on the sign language datasets.}
\label{tab:ab_perf_pretraining}
\begin{tabular}{lccc}
\toprule
\textbf{Dataset}
&  \textbf{Precision} & \textbf{Recall} & \textbf{F1-Score}  \\
\midrule
Phoenix14T  & 0.52   &   0.39 &  0.44    \\
CSL-Daily      &0.38 &   0.34  &   0.36  \\
\bottomrule
\end{tabular}
\end{table}

\paragraph{Impact of pseudo-gloss selection.} In Table \ref{tab:ab_pseudo_select}, we illustrate the impact of utilizing POS on precision, recall, and F1-score. Notably, when all words are used as tokens, recall significantly decreases from 0.39 to 0.28. This result validates our assertion that not all words have corresponding signs in sign language.

\begin{table}[ht]
\centering
\caption{Ablation of pretraining results on Phoenix14T using all words as tokens vs the selected pseudo-glosses with a threshold of 0.2}
\label{tab:ab_pseudo_select}
\begin{tabular}{lccc}
\toprule
\textbf{Tokens}
&  \textbf{Precision} & \textbf{Recall} & \textbf{F1-Score}  \\
\midrule
all words   & \textbf{0.55}   &   0.28  &  0.37   \\
pseudo-glosses      & 0.52   &   \textbf{0.39} &  \textbf{0.44}    \\
\bottomrule
\end{tabular}
\end{table}

\paragraph{Impact of Sign Encoder.} In Table \ref{tab:ab_signencoder}, we demonstrate the importance of the sign encoder on pretraining. The role of the sign encoder forms an important part of the model as it learns temporal features.

\begin{table}[ht]
\centering
\caption{Ablation of pretraining results on Phoenix14T with no sign encoder and the inclusion of a sign encoder.}
\label{tab:ab_signencoder}
\begin{tabular}{lccc}
\toprule

&  \textbf{Precision} & \textbf{Recall} & \textbf{F1-Score}  \\
\midrule
no sign encoder  & 0.50    &  0.25  &  0.33    \\
sign encoder      &\textbf{0.52}    &  \textbf{0.39}  & \textbf{0.44}    \\
\bottomrule
\end{tabular}
\end{table}

\subsection{Impact of model size.}
In Table \ref{tab:ab_size}, we demonstrate the results of our approach with the smaller language model XGLM-564M. The table shows a marginal performance reduction on the Phoenix14T dataset compared to the larger XGLM-1.7B while still outperforming GFSLT-VLP \citep{zhou2023gloss} with a similar size language model, mBART \citep{liu2020multilingual}.\\

\begin{table}[ht]
\centering
\caption{Ablation of XGLM backbones on Phoenix14T using our Sign2GPT architecture.}
\label{tab:ab_size}
\begin{tabular}{lcc}
\toprule
\multirow{2}{*}{\textbf{Backbone}} & \multicolumn{2}{c}{\textbf{Test Set}} \\
\cmidrule(lr){2-3}
&  \textbf{BLEU4} & \textbf{ROUGE}  \\
\midrule
GFSLT-VLP (mBART) \citep{zhou2023gloss}    &21.44 & 42.49 \\
\midrule
Sign2GPT(w/ PGP) (XGLM-564M)     & 22.29 & 48.21 \\
Sign2GPT(w/ PGP) (XGLM-1.7B)     & \textbf{22.52}  & \textbf{48.90} \\
\bottomrule
\end{tabular}
\end{table}

\end{document}